\documentclass{article}

\PassOptionsToPackage{numbers, compress}{natbib}

 \usepackage[preprint]{neurips_2025}

\usepackage[utf8]{inputenc} 
\usepackage[T1]{fontenc}    
\usepackage{hyperref}       
\usepackage{url}            
\usepackage{booktabs}       
\usepackage{amsfonts}       
\usepackage{nicefrac}       
\usepackage{microtype}      
\usepackage{dsfont}
\usepackage{graphicx}
\usepackage{subcaption} 
\usepackage{amsmath}
\usepackage{amssymb}

\def\eg{\emph{e.g.}}
\def\etal{{\em et al.}}

\usepackage[table]{xcolor}
\usepackage{xcolor}
\definecolor{darkgreen}{rgb}{0.0,0.5,0.0}

\definecolor{c1}{HTML}{ffe6e6}
\definecolor{c2}{HTML}{e6e6ff}

\newcommand{\equal}[1]{{\hypersetup{linkcolor=black}\thanks{#1}}}

\title{CorBenchX: Large-Scale Chest X-Ray Error Dataset and Vision–Language Model Benchmark \\for Report Error Correction}

\author{
  Jing Zou$^1$\equal{These authors contributed equally}\ , Qingqiu Li$^2$\footnotemark[1] , Chenyu Lian$^1$, Lihao Liu$^3$, Xiaohan Yan$^4$, Shujun Wang$^{5,6}$, Jing Qin$^1$ \\
  $^1$Center for Smart Health, The Hong Kong Polytechnic University \\
  $^2$School of Computer Science, Fudan University\quad
  $^3$Amazon \quad
  $^4$Tongji University \\
  $^5$Department of Biomedical Engineering, The Hong Kong Polytechnic University \\
  $^6$Research Institute for Smart Ageing, The Hong Kong Polytechnic University 
}

\begin{document}

\maketitle

\begin{abstract}
AI-driven models have shown great promise in detecting errors in radiology reports, yet the field lacks a unified benchmark for rigorous evaluation of error detection and further correction. To address this gap, we introduce \textbf{CorBenchX}, a comprehensive suite for automated error detection and correction in chest X-ray reports, designed to advance AI-assisted quality control in clinical practice. We first synthesize a large-scale dataset of 26,326 chest X-ray error reports by injecting clinically common errors via prompting DeepSeek-R1, with each corrupted report paired with its original text, error type, and human-readable description. Leveraging this dataset, we benchmark both open- and closed-source vision–language models (\eg, InternVL, Qwen-VL, GPT-4o, o4-mini, and Claude-3.7) for error detection and correction under zero-shot prompting. Among these models, o4-mini achieves the best performance, with 50.6 \% detection accuracy and correction scores of BLEU 0.853, ROUGE 0.924, BERTScore 0.981, SembScore 0.865, and CheXbertF1 0.954, remaining below clinical-level accuracy, highlighting the challenge of precise report correction. To advance the state of the art, we propose a multi-step reinforcement learning (MSRL) framework that optimizes a multi-objective reward combining format compliance, error-type accuracy, and BLEU similarity. We apply MSRL to QwenVL2.5-7B, the top open-source model in our benchmark, achieving an improvement of 38.3\% in single-error detection precision and 5.2\% in single-error correction over the zero-shot baseline. 
\end{abstract}

\section{Introduction}
\label{sec:intro}

\begin{figure}
  \centering
  \includegraphics[width=\linewidth]{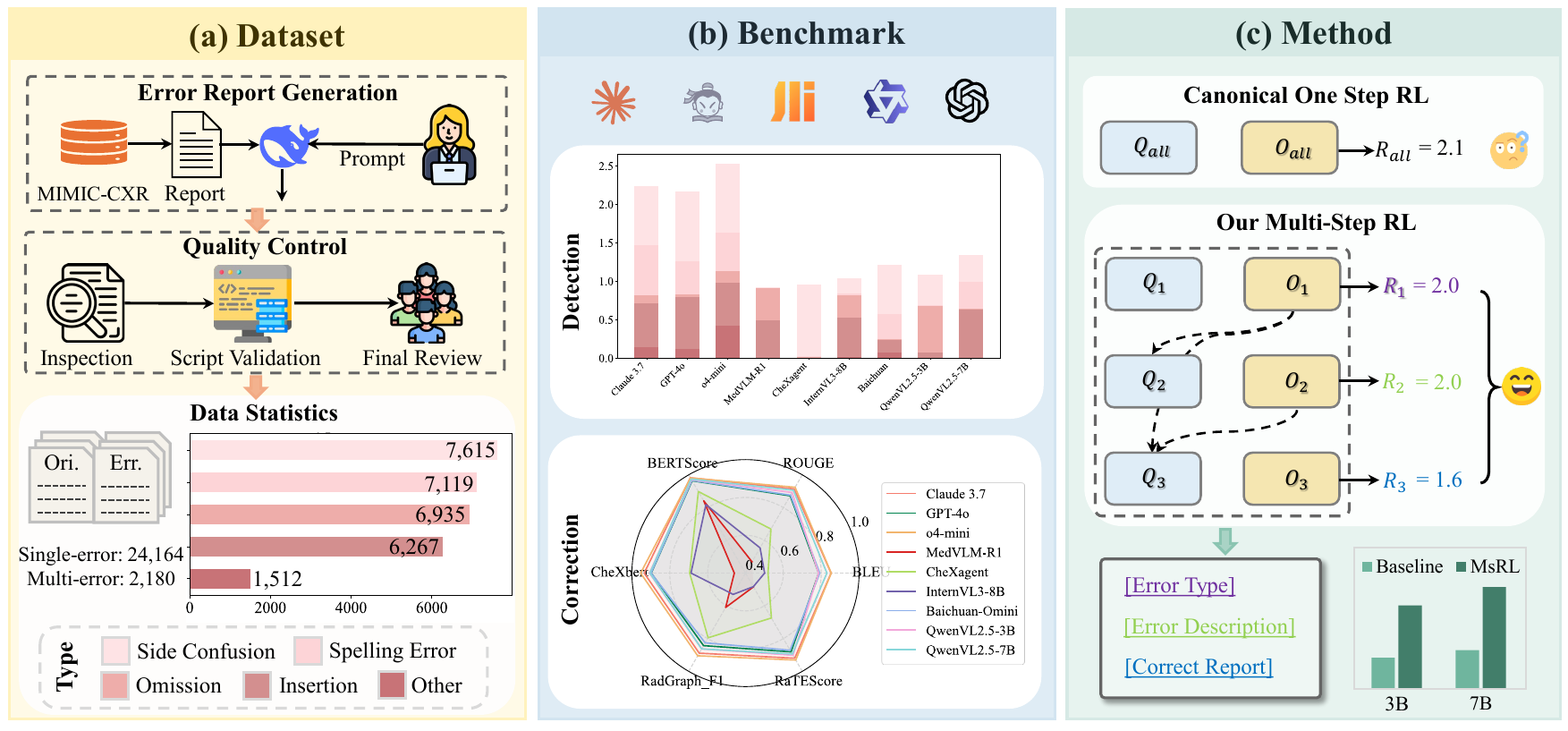}
  \caption{\textbf{Overview of CorBenchX.} (a): Error report dataset construction pipeline and dataset statistics. (b): Benchmark results across nine vision–language models for error detection and correction. (c): Illustration of our proposed multi-step reinforcement learning (MsRL) method and its performance improvements over the baseline.}
  \label{fig_overview}
\end{figure}

In modern clinical practice, radiology examination is indispensable, and the demands are increasing due to aging populations, broader imaging recommendations in updated clinical guidelines, and the increasing availability of equipment \cite{afshari2025growing}. 
As demand surges, radiologists face escalating workloads, which in turn heightens the risk of diagnostic errors in radiology reports~\cite{kim2025large}. 
%
%
%
%
To alleviate diagnostic errors in radiology reports, general healthcare systems employ a two-tiered reporting workflow: resident physicians draft preliminary reports that are reviewed, corrected, and finalized by board-certified radiologists~\cite{gertz2024potential}. While this hierarchical process improves accuracy, it demands extensive human resources and is time-consuming.
Despite such efforts, diagnostic errors, including misdiagnoses, missed diagnoses, and delayed diagnoses, occur at rates as high as 10–26\%~\cite{zhang2023diagnostic,pesapane2024errors}. These errors not only pose serious threats to patient safety and impose substantial economic burden but also increase the likelihood of malpractice suits against radiologists~\cite{kasalak2023work}.

Given these persistent challenges, there is increasing interest in leveraging Large language Models (LLMs) to streamline radiology reporting and reduce human burden—yet current approaches face critical limitations. 
Recent advances in LLMs have catalyzed interest in automated radiology report generation~\cite{chen2024chexagent,lee2023llm,chen2023cross,tanno2025collaboration,lang2025dacg}. 
LLM-driven systems can draft impressions and suggest follow-up recommendations, promising to alleviate radiologists' workload. However, despite their fluency, these generative approaches often fall short of clinical-grade reliability. Common issues such as hallucinated findings, formatting inconsistencies, and domain-mistranslations remain prevalent \cite{zeng2024enhancing}, necessitating extensive human oversight and limiting their integration into real-world clinical workflows.
\textit{In contrast to the majority of prior work on generating reports, we shift the emphasis toward automated error detection and correction in radiology reports, which is a critical yet underexplored task in radiology AI. }

Recent studies have demonstrated the potential of LLMs for automated error detection in radiology reports \cite{gertz2024potential,kim2025large,salam2025large,yan2025use}, and several specialized error dataset have been introduced, such as ReXVal~\cite{yu2023radiology} and RadEvalX~\cite{calamidaradiology}, RRED~\cite{min2022rred}, and ReXErr~\cite{rao2024rexerr}. However, these efforts exhibit critical limitations: 1) most evaluations rely on small, manually curated corpora that fail to represent the full diversity of clinical reporting mistakes; 2) they focus exclusively on error detection, offering no end‐to‐end correction; and 3) many datasets are either not publicly accessible or omit clinically common error types such as laterality confusion. Moreover, there is currently no unified benchmark that evaluates both detection and correction across a large-scale, systematically constructed dataset. 
%

To address these gaps, we introduce CorBenchX, a comprehensive benchmark for error detection and correction in chest X-ray reports.
As illustrated in Figure \ref{fig_overview}, we first construct a novel and large-scale chest X-ray error dataset derived from the MIMIC-CXR dataset \cite{johnson2019mimic} by injecting clinically common mistakes via DeepSeek-R1 prompting. Then we rigorously benchmark nine open- and closed-source VLMs for error detection and correction under zero-shot prompting. Finally, we propose a multi-step reinforcement learning method that optimizes for format compliance, error-type accuracy, and textual fidelity, yielding substantial improvements (38.3\% in detection and 5.2\% in error correction) over the baseline model.
To sum up, our contributions are threefold:
\begin{itemize}
    \item We present CorBenchX, a large-scale dataset comprising 26,326 chest X-ray error reports, including 24,146 single-error and 2,180 multi-error cases, each annotated with error spans, error type, and concise descriptions.
    \item We conduct extensive evaluations on the error dataset with various open and closed-source VLMs for both single-error and multi-error detection and correction. The results reveal that current VLMs, while powerful, fall short of meeting the clinical precision required for reliable error detection and correction in radiology reports.
    \item We propose a novel multi-step reinforcement learning framework to enhance the VLMs via sequential reasoning for error detection, description, and correction.
\end{itemize}
\section{Related Works}

\subsection{Radiology Report Generation and Evaluation}
Automated radiology report generation has rapidly evolved through several methodological paradigms. Early encoder–decoder frameworks combined convolutional or transformer-based image encoders with BERT-style decoders to directly translate image features into narrative reports \cite{syeda2020chest,wang2022automated}. 
Retrieval-based approaches, such as MedWriter~\cite{yang2021writing}, which incorporated a hierarchical retrieval mechanism and a hierarchical-LSTM decoder
to generate the report by fusing the features from the previous modules.
CXR-RePaiR \cite{endo2021retrieval}, leverage pre-trained contrastive image–text embeddings to retrieve the most similar reports from a large corpus and adapt them to new cases.  
More recently, large-scale vision–language pretraining has enabled great progress in the automatic report generation, such as CheXagent \cite{chen2024chexagent}, LLM-CXR \cite{lee2023llm}, and VLCI\_MIMIC \cite{chen2023cross}. 
Furthermore, ReXrank \cite{zhang2024rexrank} provides a public leaderboard for report generation evaluation, where 8 metrics 
are adopted as evaluation metrics.

Previous evaluation of generated reports has largely depended on lexical similarity, such as ROUGE-L~\cite{lin2004rouge} and BLEU~\cite{papineni2002bleu}, which often fail to capture subtle, clinically meaningful edits. 
To address this, medical-entity–centric measures have been proposed: CheXbert F1 \cite{smit2020chexbert} assesses agreement in disease labels inferred from text, while RadGraph-F1 \cite{yu2023evaluating} evaluates the accuracy of extracted entity–relation graphs that encode findings and anatomical locations. 
More recently, LLM-related metrics like GREEN \cite{ostmeier2024green} leverage LLM for error annotation, yielding both quantitative scores and qualitative explanations of clinically significant mistakes. 

\subsection{Report Error Detection}
L(V)LMs have recently been applied to detecting errors in radiology reports. 
Gertz \etal \cite{gertz2024potential}  evaluated GPT-4 on 100 chest X‑ray reports with synthetically introduced errors, reporting an average detection accuracy of 82.7\%, which surpassed radiology residents (80.0\%) but remained below senior radiologists.
Similarly, Kim \etal \cite{kim2025large} injected interpretive and factual errors into 300 reports, finding that GPT-4 achieved 84\% accuracy on interpretive errors and 89\% on factual errors.
Salam \etal \cite{salam2025large} evaluated open-source (Llama 3‑70B, Mixtral 8x22B) and closed-source (GPT-4o) models, with GPT-4o significantly outperforming others.
Yan \etal \cite{yan2025use} extended error detection to Chinese ultrasound reports, evaluating 400 reports containing 243 annotated errors; Claude 3.5 Sonnet achieved the highest detection rate (52.3\%).
Although these studies underscore the potential of L(V)LMs for automated report review, they exhibit key limitations: existing works rely on small, manually curated error sets that may not capture the full spectrum of clinically observed errors; most works focus solely on error detection without providing automated correction or revision, limiting their practical utility; few methods leverage multi-modal context by incorporating the original radiographic images alongside textual reports, thereby overlooking valuable cross-modal cues; and many approaches depend on human-in-the-loop validation, which restricts scalability in high-throughput clinical environments.
%

\subsection{Error Detection Dataset}
Several datasets have been introduced for radiology report error detection. 
Yu \etal \cite{yu2023radiology} proposed the ReXVal dataset, which includes 200 AI-generated/ground-truth report pairs that six radiologists evaluated for clinically significant versus insignificant errors. 
RadEvalX \cite{calamidaradiology} comprising 74 chest X-ray reports generated by an M2Tr model on IU-Xray cases, each meticulously annotated by expert radiologists for the presence and clinical severity of reporting. 
RRED \cite{min2022rred} utilized a "generator" to generate findings-impression inconsistent errors in MIMIC-CXR reports and supplemented this with manual error annotations by two radiologists on 111 cases. 
Sun \etal \cite{sun2025generative} generated 1,656 chest X-ray reports using GPT-4. Half were error-free; the other half contained errors introduced via prompts. Meanwhile, an additional set of 307 real MIMIC-CXR reports was paired with 307 GPT-4 versions containing errors. 
While these datasets offer valuable insights into error analysis and detection, their small scale (no more than 200 cases in ReXVal and RadEvalX) and limited public accessibility (RRED and Sun's dataset) hinder their suitability for large-scale evaluation.
ReXErr \cite{rao2024rexerr} delivers a public large-scale dataset for chest X-Ray error detection. However, its uniform injection of exactly three errors per report fails to mirror real-world error distributions, omits critical categories such as laterality confusion, and risks introducing internally contradictory mistakes. Moreover, ReXErr does not include standardized error detection and correction benchmarks.

\section{Error Report Dataset Construction}
\label{sec:dataset}

\begin{figure}
  \centering
  \includegraphics[width=\linewidth]{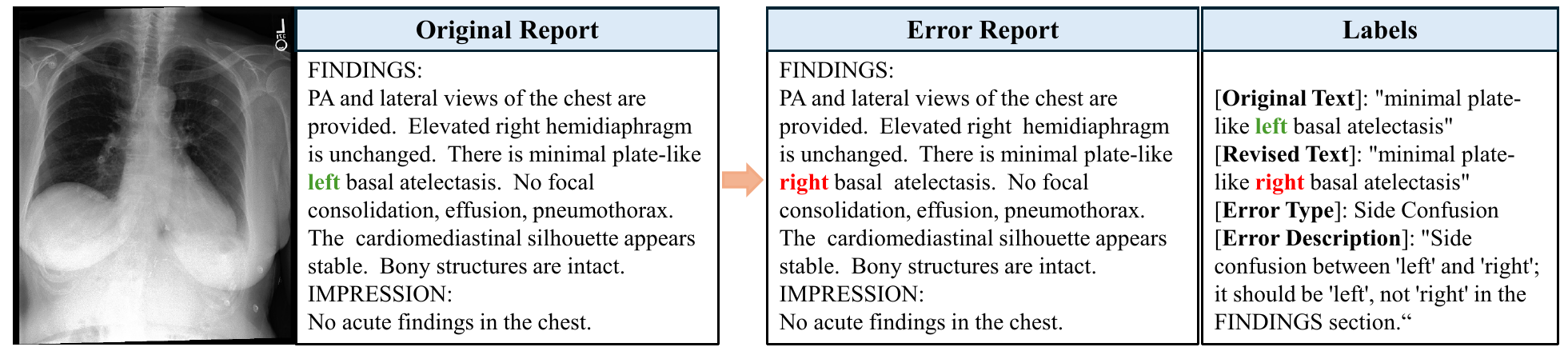}
  \caption{Example of a chest X-ray, paired original radiology report, and the corresponding error-injected report with labels. Text spans highlighted in \textcolor{red}{red} denote the injected errors, while the corrected spans are shown in \textcolor{darkgreen}{green}.}
  \label{report}
\end{figure}

\begin{table} \small
  \caption{Error type explanation and data statistics}
  \label{sample-table}
  \centering
  \begin{tabular}{llr}
    \toprule
    Error type   & Description  & Number     \\
    \midrule
    Omission    & Missing relevant clinical findings or words   &6,267  \\
    Insertion     &The unintentional insertion of incorrect words or expressions   & 6,935   \\
    Spelling Error     &Spelling mistakes or typos     &7,119\\
    Side Confusion  &Errors involving side or orientation    &7,615\\
    Other   &Mistakes in units of measurement, punctuation mistakes, etc.   &1,512 \\
    Total   & -  &29,448 \\
    \bottomrule
  \end{tabular}
\label{tab_errortype}
\end{table}

We introduce \textbf{CorBenchX}, a high-quality and systematically constructed dataset for chest X-ray report error detection and correction. The dataset simulates realistic reporting errors across a range of clinically motivated categories, providing a reliable foundation for training and evaluation.

\textbf{Dataset Source and Sampling.} 
CorBenchX is built on the publicly available MIMIC-CXR dataset~\cite{johnson2019mimic}, which contains de-identified chest X-ray reports collected from Beth Israel Deaconess Medical Center. We extract the ``Findings'' and ``Impression'' sections from each report and remove records where both sections are empty. From the resulting pool, we randomly sample 26,326 clean reports as the basis for synthetic error injection.

\textbf{Error Injection Procedure.} 
To create realistic errorful variants, we use the DeepSeek-R1 API with a carefully designed prompt (see Appendix for details). Each API call outputs an error-injected report along with: (i) the error type label, (ii) the paired original and altered text spans, and (iii) a concise natural language error description (see Figure~\ref{report}). 
To simulate clinically relevant reporting errors, we introduce structured perturbations into clean reports
in two types of samples:
\begin{itemize}
    \item \textbf{Single-error reports:} Each report contains exactly one error from one of five categories—\textit{omission}, \textit{insertion}, \textit{spelling error}, \textit{side confusion}, or \textit{other}—resulting in 24,146 single corrupted samples.
    \item \textbf{Multi-error reports:} To better reflect real-world reporting complexity, we additionally generate 2,180 reports containing two to three independent errors.
\end{itemize}

\textbf{Quality Control Pipeline.} 
To ensure high-quality annotations, we implement a three-stage quality control process (Figure~\ref{fig_overview} (a)).  \textbf{Stage 1: Expert Inspection.} Human annotators examine all reports to flag and correct misplaced or implausible error insertions.
\textbf{Stage 2: Script Validation.} Automated scripts validate formatting consistency, detect unchanged or malformed edits, remove redundant symbols, and ensure that exactly one or the intended number of edits exist per sample.
\textbf{Stage 3: Final Review.} Annotators conduct a second pass to resolve edge cases and correct any ambiguous or residual inconsistencies that may have passed earlier filters.
This combined human–in-the-loop pipeline ensures both scalability and reliability in dataset construction.

\textbf{Dataset Composition and Availability.}
The final version of CorBenchX consists of clean–corrupted report pairs with detailed annotations, including error type, span-level edits, and error descriptions. An example of a CXR image and its associated reports is shown in Figure~\ref{report}, while Table~\ref{tab_errortype} summarizes the error categories and their distributions. The dataset serves as a comprehensive benchmark for developing and evaluating radiology report error detection and correction systems.
The complete dataset has been submitted to PhysioNet and is currently under review; it will be publicly available soon.

\section{Multi-Step Reinforcement Learning}
\label{sec:method}
Correcting radiology report errors requires precise localization of erroneous spans and flexible, context-aware revision strategies. Due to the diverse linguistic patterns across error types, fixed or templated supervision is often inadequate. To address this, we introduce a novel method and formulate the task as a {three-stage} reinforcement learning problem that promotes step-by-step reasoning and fine-grained correction. We adopt {Group Relative Policy Optimization (GRPO)} as the training objective to guide the model toward clinically consistent and contextually appropriate revisions.

\subsection{Three Stage Reinforcement Learning Optimization}

\begin{figure}[!t]
  \centering
  \includegraphics[width=\linewidth]{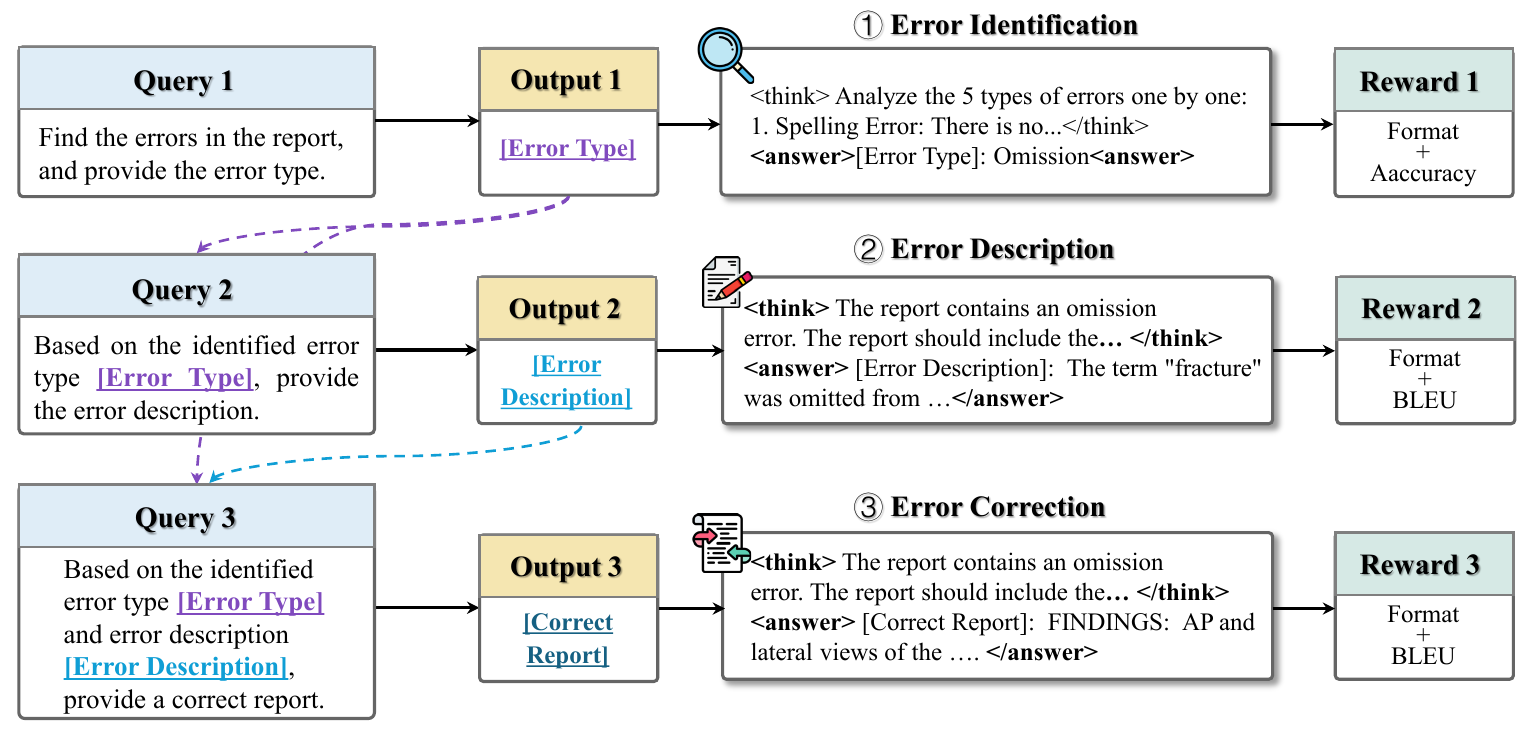}
  \caption{Illustration of our multi-step reinforcement-learning framework: the model sequentially performs error identification, description, and correction, with each stage guided by a tailored reward.}
  \label{fig:method}
\end{figure}

The report correction task can be decomposed into three stages: \textbf{$ \text{error identification} \rightarrow \text{error description} \rightarrow \text{error correction} $}. Based on this, we design a multi-step approach that breaks the complete trajectory into multiple sub-trajectories to encourage the model to perform clear and targeted reasoning at each step, thereby enabling supervision over intermediate reasoning processes, as illustrated in Figure \ref{fig:method}. Formally, the reasoning trajectory is denoted as
\begin{equation}
\mathcal{T}  = ((Q_1, O_1), …, (Q_K, O_K)),
\end{equation}
where $Q_k$ and $O_k$ denote the model’s query and output at each step, respectively. $K$ represents the total number of steps required by the reasoning trajectory, which is set to 3 in our task. The first state $Q_1$ serves as the initial prompt. Each subsequent query $Q_k$ contains the previous query $Q_{k-1}$ and the corresponding output $O_{k-1}$. 
%


\textbf{Step 1: Error Identification.}
First, we supervise the model to correctly identify the error type by optimizing classification accuracy. The reward for this step, denoted as $R_1$ , is the sum of the Format Reward and the Accuracy Reward.
\textbf{Format Reward:} The format reward $R_{format} \in \{0,1\}$ is designed to ensure that the model encloses its reasoning within the designated tags (e.g., <think> and </think>) and wraps the final answer within <answer> and </answer> tags.
\begin{equation}
R_{format} = \mathds{1}(match(content)),
\end{equation}
where $match$ denotes the regular expression matching operation.

\textbf{Accuracy Reward:} The accuracy reward $R_{acc} \in \{0,1\}$ is set to 1 if the model correctly identifies the current error type, and 0 otherwise.
\begin{equation}
R_{acc} = \mathds{1}(Err_{pred}=Err_{gt}),
\end{equation}
where $Err_{pred}$ denotes the model's predicted error type, and $Err_{gt}$ refers to the ground truth.

\textbf{Step 2: Error Description.} Based on the Step 1, we perform error description to help the model better understand and localize different types of errors. This step also enables the model to provide users with more detailed references and explanations during interaction. We supervise the quality of error description using the Format Reward and the BLEU Reward. The Format Reward is the same as above, and the \textbf{BLEU Reward} is defined as follows.
\begin{equation}
R_{bleu} = BLEU(Des_{pred}, Des_{gt}),
\end{equation}
where $Des_{pred}$ denotes the model's predicted description, and $Des_{gt}$ refers to the ground truth.

\textbf{Step 3: Error Correction.}
Building on the previous two steps, the model conducts evidence-based error correction, with the accuracy of the corrections supervised by the Format and BLEU Reward.

\subsection{Training with GRPO}
\label{sec: grpo}
The model’s policy is optimized to maximize the cumulative reward over the entire trajectory for 3 stages RL learning, formulated as: 
\begin{equation}
J(\theta) = \sum_{k=1}^{K} J^k(\theta).
\end{equation}
Here, $\pi_{\theta}$ is the policy model parametrized by $\theta$. $J^k(\theta)$ denotes the optimization objective at step $k$.
We employ GRPO~\cite{deepseek}, a variant of PPO~\cite{ppo} that introduces advantage normalization within grouped samples, as the optimization objective at each step. This objective guides the policy to generate structurally coherent and instruction-following report error corrections.
\begin{equation}
\begin{aligned}
J^k(\theta) =\; & \mathbb{E}[q^k \sim P(Q^k),\, \{o_i^k\}_{i=1}^G \sim \pi_{\theta_{\text{old}}}(O^k | q^k)]  \frac{1}{G} \sum_{i=1}^G \bigg( \min\Big( 
\frac{\pi_{\theta}(o_i^k | q^k)}{\pi_{\theta_{\text{old}}}(o_i^k | q^k)} A_i^k, \\
& \text{clip}\big( 
\frac{\pi_{\theta}(o_i^k | q^k)}{\pi_{\theta_{\text{old}}}(o_i^k | q^k)},\, 
1 - \varepsilon,\, 1 + \varepsilon \big) A_i^k \Big) 
- \beta D_{\text{KL}}(\pi_{\theta} \| \pi_{\text{ref}}) \bigg). 
\end{aligned}
\end{equation}
where $\pi_{\theta_{\text{old}}}$ presents the old policy model, $Q_k$ is the query for step $k$, $\varepsilon$ and $\beta$ are hyperparameters, G denotes the number of outputs within a group. $A_i^k$ is the advantage calculated based on rthe relative rewards of the outputs within each group. During training, the number of grouped samples is set to 8.

\section{Evaluation}
\label{sec:benchmark}


\subsection{Experimental Settings}
\textbf{Evaluation Models.}
We evaluate nine vision-language models (VLMs) alongside our proposed method under a zero-shot setting for two tasks: error detection and error correction in chest X-ray reports. The evaluated models include six open-source VLMs: MedVLM-R1~\cite{pan2025medvlm}, CheXagent~\cite{chen2024chexagent}, InternVL3-8B~\cite{zhu2025internvl3}, Baichuan-Omini-1.5-7B~\cite{li2025baichuan}, QwenVL2.5-3B, and QwenVL2.5-7B~\cite{bai2025qwen2}; and three closed-source models: Claude 3.7 Sonnet, GPT-4o~\cite{achiam2023gpt}, and o4-mini~\cite{openai-o4-mini-2025}.


\textbf{Implementation Details.}
All experiments are conducted on NVIDIA A800 GPUs.
For each model, we prompt it to perform two tasks in sequence: (1) identify and classify the error type in a corrupted report, and (2) generate a corrected version of the report. No additional fine-tuning or in-domain training is performed. 
Detailed hyperparameters and prompting templates are provided in the Appendix.


\begin{figure}[!t]
  \centering
  \includegraphics[width=\linewidth]{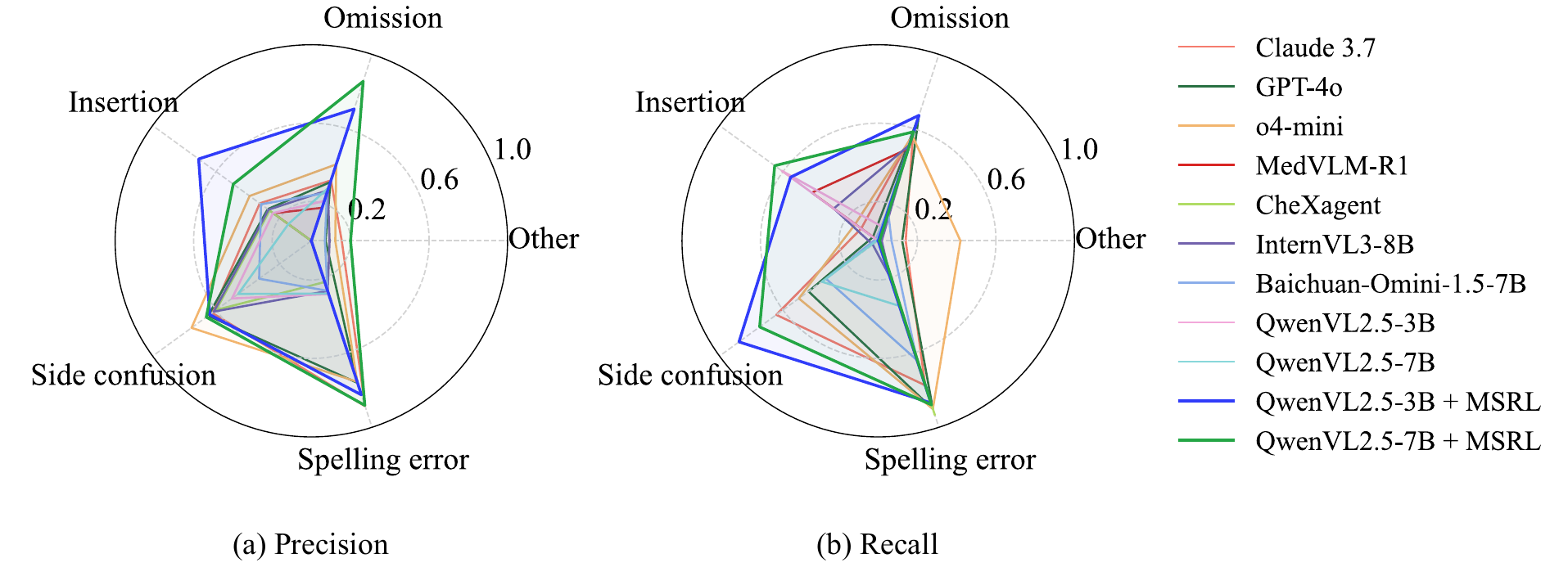}
  \caption{Precision and recall for single-error detection across various VLMs and models enhanced by our MSRL, broken down by the five error categories.}
  \label{fig:detect_single}
\end{figure}

\begin{table} \small
  \caption{Evaluation results on single-error report correction (report-level)/(\colorbox{gray!15}{sentence-level}). The highest score in each column is highlighted in \colorbox{c1}{pink}, and the second-best in \colorbox{c2}{blue}.}
  \label{sample-table}
  \centering
  \begin{tabular}{l|ccccccc}
    \toprule
    \textbf{Model}   &BLEU &ROUGE &BERTScore &SembScore &CheXbertF1 &RadGraphF1     \\
    \midrule
    Claude 3.7 sonnet   &0.852   &0.914 &0.982 &0.817 &0.935 &0.889\\
    GPT-4o   &0.787   &0.872 &0.964 &0.782 &0.898 &0.843\\
    o4-mini   &0.853   &0.924 &0.981 &0.865 &0.954 &0.905\\
    \midrule
    MedVLM-R1    &0.315   &0.469  &0.841 &0.459 &0.610 &0.484\\
    CheXagent     &0.519   &0.669  &0.898  &0.695 &0.795 &0.674\\
    InternVL3-8B   &0.768   &0.848 &0.948 &0.777 &0.903 &0.813\\
    Baichuan-Omini1.5-7B   &0.792   &0.876 &0.966 &0.784 &0.899 &0.826\\
    QwenVL2.5-3B   &0.786    &0.892 &0.971 &0.807 &0.907 &0.863\\
    QwenVL2.5-7B      &0.830  &0.906 &0.974 &0.793 &0.905 &0.863\\
    \midrule
    \textbf{QwenVL2.5-3B+MSRL}   &   \cellcolor{c2}{0.938}  &\cellcolor{c2}{0.971} &\cellcolor{c2}{0.993} &\cellcolor{c2}{0.839} &\cellcolor{c2}{0.951} &\cellcolor{c2}{0.931}\\
   \textbf{QwenVL2.5-7B+MSRL}      & \cellcolor{c1}{0.960} & \cellcolor{c1}{0.984} &\cellcolor{c1}{0.997} &\cellcolor{c1}{0.905} &\cellcolor{c1}{0.984} &\cellcolor{c1}{0.958} \\
    \bottomrule
    \toprule
    \midrule
     \rowcolor{gray!15}{Claude 3.7 sonnet}&0.345   &0.477 &0.862 &0.789 &0.815 &0.416\\
    \rowcolor{gray!15}{GPT-4o}          &0.365   &0.550 &0.870 &\cellcolor{c2}{0.795} &0.843 &0.465\\
    \rowcolor{gray!15}{o4-mini}         &0.386   &0.547 &0.876 &0.852 &\cellcolor{c2}{0.878} &0.482\\
    \midrule
    \rowcolor{gray!15}{MedVLM-R1}       &0.282   &0.441  &0.826 &0.508 &0.646 &0.406\\
    \rowcolor{gray!15}{CheXagent}      &0.326   &0.481  &0.840  &0.665 &0.706 &0.413\\
    \rowcolor{gray!15}{InternVL3-8B}   &\cellcolor{c1}{0.516}   &\cellcolor{c1}{0.719} &0.914 &0.775 &0.863 &{0.606}\\
    \rowcolor{gray!15}{Baichuan-Omini1.5-7B}&\cellcolor{c2}{0.504}&\cellcolor{c2}{0.713} &\cellcolor{c2}{0.920} &0.762 &0.862 &\cellcolor{c1}{0.591}\\
    \rowcolor{gray!15}{QwenVL2.5-3B}   &0.486    &0.702 &0.919 &0.773 &0.836 &0.580\\
   \rowcolor{gray!15}{QwenVL2.5-7B}   &0.467  &0.686 &0.911 &0.790 &0.849 &0.554\\
    \midrule
     \rowcolor{gray!15}{\textbf{QwenVL2.5-3B+MSRL}}   &   {0.481}  & {0.701}  & \cellcolor{c1}{0.921}  & {0.764}  & {0.824}  & \cellcolor{c2}{0.558} \\
    \rowcolor{gray!15}{\textbf{QwenVL2.5-7B+MSRL}}      & {0.400}  & {0.536}  & {0.868}  & \cellcolor{c1}{0.897}  & \cellcolor{c1}{0.929}  & {0.446} \\   
    \bottomrule
  \end{tabular}
\label{tab_language_single}
\end{table}

\textbf{Evaluation Metrics.} 
We assess each model’s performance along three dimensions: 
(1) Error detection: measured by precision and recall over the five error types; (2) Error correction in report level: assessed with two word level metrics: BLEU \cite{papineni2002bleu} and ROUGE \cite{lin2004rouge}, two semantic level metrics: BERTScore \cite{zhang2019bertscore} and SembScore \cite{smit2020chexbert}, and two clinical efficacy level metrics: CheXbert \cite{smit2020chexbert} and RadGraph-F1 \cite{yu2023evaluating}; and (3) Error correction in sentence level: apply the same suite of six metrics to the individual corrected sentences, enabling fine‐grained assessment of local edits.

\subsection{Experimental Results and Analysis}
We first evaluate the performance of nine baseline VLMs on both single-error and multi-error detection and correction tasks. We then compare these results with our proposed MSRL-enhanced models—\textbf{QwenVL2.5-3B+MSRL} and \textbf{QwenVL2.5-7B+MSRL}—to assess the effectiveness of multi-step reinforcement learning in improving fine-grained clinical reasoning and radiology report correction.
Finally, we conduct an ablation study to validate the contribution of our multi-step RL framework compared to standard single-step reinforcement learning.

\textbf{Results on Single-error Detection and Correction.} \quad  
Figures~\ref{fig:detect_single} (a) and \ref{fig:detect_single} (b) present per-error-type precision and recall for single-error detection across nine evaluated vision-language models (VLMs).
Table~\ref{tab_language_single} summarizes the corresponding error correction performance, evaluated using six metrics at both the report (upper part) and sentence levels (lower part).
 As shown in Figure~\ref{fig:detect_single}, \textbf{o4-mini} achieves the best overall detection performance, with an average precision of 0.486 and recall of 0.506. In terms of correction quality (Table~\ref{tab_language_single}), closed-source models—Claude 3.7 Sonnet, GPT-4o, and o4-mini—consistently outperform their open-source counterparts in report-level metrics, with o4-mini ranking highest across all evaluation scores.
Within open‐source models, \textbf{QwenVL2.5-7B} leads the pack,
whereas MedVLM-R1 performs markedly worse.
%
Generally, sentence‐level metrics (lower part) are substantially lower than report‐level scores, demonstrating that localized, span‐level evaluation reveals challenges masked by full‐report metrics. 
Across all models, current error correction capabilities of existing VLMs fall short of clinical-grade reliability, reinforcing the need for more targeted and interpretable strategies.

%

\begin{figure}[!t]
  \centering
  \includegraphics[width=\linewidth]{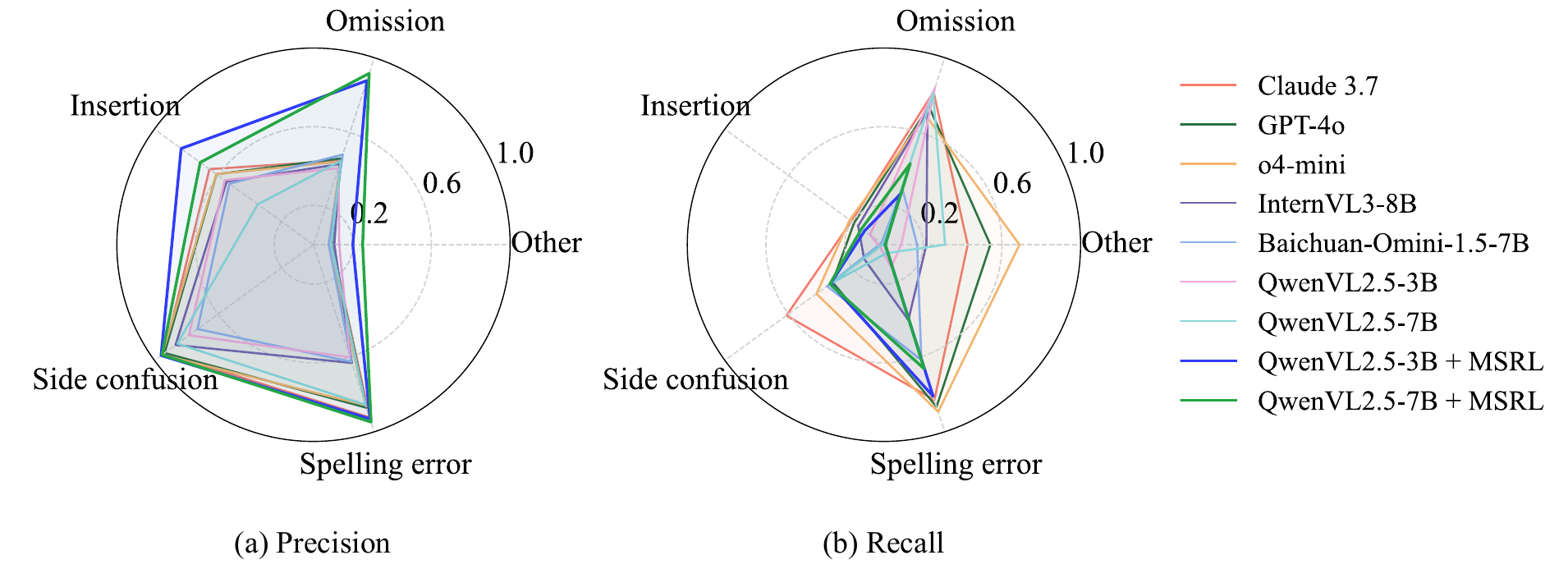}
  \caption{Precision and recall for multi-error detection across various VLMs and models enhanced by our MSRL, broken down by the five error categories.}
  \label{fig:detect_multi}
\end{figure}

\textbf{Results on Multi-error Detection and Correction.}
Figure \ref{fig:detect_multi} (a) and Figure \ref{fig:detect_multi} (b) depict per‐error‐type precision and recall for multi‐error detection across all evaluated VLMs. Closed‐source models again dominate: Claude 3.7 achieves the highest average precision (0.612), while o4-mini attains the highest average recall (0.580), both substantially outperforming open‐source models. 
Table~\ref{tab_language_multi} reports multi‐error correction performance under the same six metrics. 
At the report level, QwenVL2.5-3B  is the top open‐source performer.
The results are far lower than those for single-error correction, underscoring the substantial challenge that multi‐error correction poses for current VLMs.
At the sentence level, Baichuan-Omini1.5-7B obtains the best results.
%
Notably, o4-mini underperforms because it paraphrases entire reports instead of making focused span‐level corrections.

\begin{table} \small
  \caption{Evaluation results on multi-error report correction (report-level)/(\colorbox{gray!15}{sentence-level}). The highest score in each column is highlighted in \colorbox{c1}{pink}, and the second-best in \colorbox{c2}{blue}.}
  \label{sample-table}
  \centering
  \resizebox{0.99\textwidth}{!}{
  \begin{tabular}{l|ccccccc}
    \toprule
    Model    &BLEU &ROUGE &BERTScore &SembScore &CheXbertF1 &RadGraphF1     \\
    \midrule
    Claude 3.7 sonnet   &0.701   &0.817 &0.959 &0.724 &0.856 &0.773\\
    GPT-4o              &0.629   &0.769 &0.935 &0.669 &0.801 &0.719\\
    o4-mini             &0.404   &0.619 &0.909 &0.670 &0.788 &0.596\\
    \midrule
    InternVL3-8B        &0.685    &0.808 &0.940 &0.682 &0.811 &0.748\\
    Baichuan-Omini1.5-7B &0.755   &0.875 &0.966 &0.753 &0.876 &0.817\\
    QwenVL2.5-3B        &0.742    &0.859 &0.964 &0.736 &0.855 &0.805\\
    QwenVL2.5-7B        &0.728   &0.847 &0.959 &0.712 &0.833 &0.780\\
    \midrule
    \textbf{QwenVL2.5-3B+MSRL}   &   \cellcolor{c2}{0.874}  &\cellcolor{c2}{0.940} &\cellcolor{c2}{0.985} &\cellcolor{c2}{0.794} &\cellcolor{c2}{0.908} &\cellcolor{c2}{0.866}\\
    \textbf{QwenVL2.5-7B+MSRL}      & \cellcolor{c1}{0.900} &\cellcolor{c1}{0.958} &\cellcolor{c1}{0.992} &\cellcolor{c1}{0.852} &\cellcolor{c1}{0.948} &\cellcolor{c1}{0.898} \\
    \bottomrule
    \toprule
    \rowcolor{gray!15}{Claude 3.7 sonnet}   &0.461   &0.666 &0.919 &0.684 &0.781 &0.577\\
    \rowcolor{gray!15}{GPT-4o}              &0.502   &0.735 &0.925 &0.712 &0.807 &0.593\\
    \rowcolor{gray!15}{o4-mini}             &0.297   &0.575 &0.893 &0.724 &0.813 &0.505\\
    \midrule
    \rowcolor{gray!15}{InternVL3-8B}        &0.538    &0.761 &0.932 &0.697 &0.812 &0.631\\
    \rowcolor{gray!15}{Baichuan-Omini1.5-7B} &0.591   &0.810 &0.952 &0.737 &0.828 &0.680\\
    \rowcolor{gray!15}{QwenVL2.5-3B}        &0.569    &0.783 &0.943 &0.709 &0.810 &0.657\\
    \rowcolor{gray!15}{QwenVL2.5-7B}        &0.560   &0.800 &0.945 &0.714 &0.806 &0.640\\
    \midrule
    \rowcolor{gray!15}{\textbf{QwenVL2.5-3B+MSRL}}   &   \cellcolor{c1}{0.647}  & \cellcolor{c1}{0.848}  & \cellcolor{c1}{0.967}  & \cellcolor{c2}{0.753}  & \cellcolor{c2}{0.848}  & \cellcolor{c1}{0.693} \\
    \rowcolor{gray!15}{\textbf{QwenVL2.5-7B+MSRL}}      & \cellcolor{c2}{0.636}  & \cellcolor{c2}{0.827}  & \cellcolor{c2}{0.961}  & \cellcolor{c1}{0.829}  & \cellcolor{c1}{0.901}  & \cellcolor{c2}{0.691} \\
    \bottomrule
  \end{tabular}
  }
\label{tab_language_multi}
\end{table}

\textbf{Effectiveness of
Multi-step Reinforcement Learning.}  \quad
We perform our MSRL on Qwen-2.5-VL 3B and Qwen-2.5-VL 7B and compare its performance with other VLMs.
As shown in Figures~\ref{fig:detect_single} and Figures~\ref{fig:detect_multi}, our method achieves an average increase of 38.3\% in precision and 30.5\% in recall on the single error detection task with Qwen-2.5-VL-7B. Similarly, for multi-error detection, we observe an average improvement of 23.6\% in precision and 1.5\% in recall, validating the generalization capability of the model.
Notably, when our model is initialized with Qwen-2.5-VL 3B, its classification accuracy on the ``other'' category remains at a very low level. The underlying reason is that Qwen-2.5-VL 3B, under zero-shot settings, fails to recognize the ``other'' category and tends to ignore its analysis during the reasoning process (the content within the <think> </think>). This observation highlights that without early-stage instruction fine-tuning, RL alone yields suboptimal reasoning performance, which has been approved in \cite{deepseek, xreason}.
For report-level correction, Table~\ref{tab_language_single} shows that QwenVL2.5-3B and 7B models trained with multi-step RL outperform their zero-shot baselines by 7.4\% and 5.2\% on single-error correction. Sentence-level gains are even larger. On the more challenging multi-error task (Table~\ref{tab_language_multi}), our model improvements reach 6.8\% and 11.5\%, highlighting the effectiveness and generalization ability of our MSRL.


\textbf{Ablation Studies.}
As shown in Table \ref{tab:ablation}, we compare our MSRL with single-step RL, which incorporates all processes into a single inference and simultaneously optimizes the Accuracy Reward, Format Reward, and BLEU Reward. This approach fails to effectively follow instructions step by step, resulting in an average performance gap of 13.3\% compared to MSRL. 

\begin{table} \small
  \caption{Ablation studies on reinforcement learning (RL) and multi-step reinforcement learning (MSRL) for single-error correction.}
  \label{tab:ablation}
  \centering
  \resizebox{0.99\textwidth}{!}{
  \begin{tabular}{l|ccccccc}
    \toprule
    Model    &BLEU &ROUGE &BERTScore &SembScore &CheXbertF1 &RadGraphF1     \\
    \midrule
    RL   &0.788   &0.882 &0.944 &0.798 &0.916 &0.853\\
    MSRL   &0.938   &0.971 &0.993 &0.839 &0.951 &0.931\\
    \midrule
    RL   &0.873   &0.939 &0.978 &0.838 &0.945 &0.906\\ 
    MSRL   &0.960   &0.984 &0.997 &0.905 &0.984 &0.958\\\bottomrule
    \end{tabular}
    }
\label{tab_ablation}
\end{table}

\section{Conclusion}
\label{sec:conclusion}
In this work, we present CorBenchX, the first large-scale benchmark for automated error detection and correction in chest X-ray reports. By synthesizing 26,326 clinically motivated error cases via DeepSeek-R1, we enable a rigorous evaluation of both open- and closed-source L(V)LMs. Our experiments reveal that even the strongest models achieve just 50.6 \% error-type detection accuracy and remain below clinical-grade correction. To this end, we propose MSRL that sequentially supervises error identification, description, and correction, yieding substantial gains over the baselines.
%
%
Despite these advances, our benchmark has two main limitations. First, CorBenchX currently covers only chest X-ray data; extending to other modalities such as CT and MRI will be essential to capture the full spectrum of radiology reporting. Second, our benchmark does not address errors that arise from discrepancies with a patient’s prior imaging or disease history, for example, incorrect temporal comparisons or overlooked historical findings.
Future work will therefore focus on broadening CorBenchX to include multi‐modal datasets (CT, MRI) and incorporating electronic health record context to evaluate models’ ability to detect and correct errors related to longitudinal patient history.

\clearpage
\bibliography{Ref}
\bibliographystyle{unsrtnat}

\end{document}